\pdfoutput=1
\documentclass[11pt]{article}

\usepackage{EACL2023}
\usepackage{times}
\usepackage{latexsym}
\usepackage{soul}

\usepackage{multirow} 
\usepackage[T1]{fontenc}
\usepackage[utf8]{inputenc}

\usepackage{microtype}

\usepackage{inconsolata}

\newcommand{\ambient}{\mbox{\textsc{AmbiEnt}}}

\newcommand{\aspace}{\hspace{1em}}

\DeclareSymbolFont{extraup}{U}{zavm}{m}{n}
\DeclareMathSymbol{\vardiamond}{\mathalpha}{extraup}{87}
\newcommand{\massit}{$^{\spadesuit}$}
\newcommand{\aiTwo}{$^{\clubsuit}$}

\newcommand{\uw}{$^{\vardiamond}$}

\title{A Taxonomy of Ambiguity Types for NLP}

\author{Margaret Y. Li\uw\aspace Alisa Liu\uw\aspace Zhaofeng Wu\massit\aspace Noah A. Smith\uw\aiTwo\aspace \\
  \uw{}Paul G. Allen School of Computer Science \& Engineering, University of Washington \\
  \massit{}Massachusetts Institute of Technology \quad \aiTwo{}Allen Institute for AI \\
  \texttt{marg33@cs.washington.edu} }
\begin{document}
\maketitle

% \begin{abstract}

%  draft abstract for openreview submission

% Ambiguity is an critical component of language that allows for more effective communication between speakers, but is often ignored in NLP. Recent work suggests that NLP systems may struggle to grasp certain elements of human language understanding because they may not handle ambiguities at the level that humans naturally do in communication. Additionally, different types of ambiguity may serve different purposes and require different approaches for resolution, and we aim to investigate how language models' abilities vary across types. We propose a taxonomy of ambiguity types as seen in English to facilitate NLP analysis. Our taxonomy can help make meaningful splits in language ambiguity data, allowing for more fine-grained assessments of both datasets and model performance.

% tldr
% We propose a taxonomy of ambiguity types in language to facilitate analysis of data and model performance across types.

% \end{abstract}

\section{Introduction}

Ambiguity is a critical component of natural language that allows for more efficient communication between speakers \cite{piantadosi-etal-2012}, but is often ignored in NLP. Recent work suggests that NLP systems may struggle to grasp certain elements of human language understanding because they may not handle ambiguities at the level that humans naturally do in communication \cite{yuan-etal-2023-ambicoref, liu-etal-2023-afraid}. Additionally, different \emph{types of ambiguity} may pose different challenges as they require different approaches for resolution \cite{gibbon-2010-ambiguity}. We aim to lay the foundations for investigating how language models’ abilities and limitations vary across different ambiguity types.

In this work-in-progress, we propose a \textbf{new taxonomy of ambiguity types} seen in English to facilitate more fine-grained analysis of data and model performance. To our knowledge, no comprehensive taxonomy of ambiguity types has been created for or applied to analyzing ambiguity in the modern NLP context. \textbf{We propose a framework of eleven ambiguity types: lexical, syntactic, scopal, elliptical, collective/distributive, implicative, presuppositional, idiomatic, coreferential, generic/non-generic, and type/token.} We plan to apply this taxonomy to annotate \ambient\ \cite{liu-etal-2023-afraid}---a benchmark that captures ambiguities through entailment relations---to estimate the relative frequencies of ambiguity types, and then to make a more balanced and diverse dataset with more examples for types that are underrepresented, allowing for the development of more nuanced benchmarks. We also will analyze how models perform across different types. For the most difficult types, we hope to create targeted tasks or evaluation formats.

\section{Background}

While the creators of \ambient\ proposed seven ambiguity categories for analysis, we refine and extend it with more categories for greater coverage and more detailed explanations. Our taxonomy draws from linguistic definitions, but is not meant to satisfy all linguistic theories---as what is considered ambiguity, and how to separate different kinds of ambiguity, is a difficult problem in linguistics \cite{sennet-ambiguity-23, Walton-1996-fallacies, bach-1998-ambiguity}. However, these linguistic concerns may not be identical to those of NLP system design and evaluation. Our taxonomy marks separate ambiguity phenomena that represent different challenges in language understanding, and that are straightforward to capture in an NLP task/dataset such as natural language inference. As such, our taxonomy can help make meaningful splits in language ambiguity data, allowing for more detailed assessments of both datasets and models' abilities.

% \nascomment{maybe leave the door open explicitly for future refinement to our taxonomy?}

\section{Types of Ambiguity}

We briefly describe the categories in our taxonomy. A table of selected examples from \ambient, categorized by the authors into each ambiguity type, can be found in \autoref{sec:appendix}.

\subsection{Lexical Ambiguity}
Lexical ambiguity occurs when words have multiple possible meanings \cite{sennet-ambiguity-23, Walton-1996-fallacies}. For example, in ``\textit{We finally reached the bank},'' the word \emph{bank} could mean ``\textit{financial institution}'' or ``\textit{side of river}.'' \cite{Cruse-1986-lexical}.

\subsection{Syntactic Ambiguity}
Syntactic ambiguity happens when multiple grammatical structures are possible for a sequence of words. For instance, the phrase ``\textit{superfluous hair remover}'' could be parsed as [superfluous] [hair remover] or [superfluous hair] [remover] \cite{sennet-ambiguity-23}. For another case, consider the sentence ``\textit{The girl hit the boy with the book}.'' It is ambiguous as to whether the girl or the boy is the one with the book \cite{bach-1998-ambiguity}. 

% Syntactic ambiguity happens when multiple grammatical structures are possible for a sequence of words. This can happen within a phrase: the sentence ``superfluous hair remover'' could be parsed as [superfluous] [hair remover] or [superfluous hair] [remover] \cite{sennet-ambiguity-23}. Attachment may also be ambiguous: the sentence ``The girl hit the boy with the book'' is ambiguous as to whether the girl or the boy has the book \cite{bach-1998-ambiguity}.

\subsection{Scopal Ambiguity}
When a sentence contains multiple quantifiers or scopal expressions, their relative ordering may be ambiguous \cite{kearns-2000-semantics, Kroeger-2019, sennet-ambiguity-23}. For the sentence ``\textit{every student read two poems},'' when \textit{every} takes scope over \textit{two}, it is possible that every student read different poems. But when \textit{two} takes scope over \textit{every}, there must be two specific poems that every student read. 

\subsection{Elliptical Ambiguity}
In some sentences containing ellipsis, the identity of elided words or phrases may be ambiguous \cite{sennet-ambiguity-23}. For example, the sentence ``\textit{Peter walked his dog, and Dan did, too}'' may mean that Dan walked his own dog, or that Dan walked Peter’s dog \cite{Tomioka-1999-sloppy}. Similarly, the sentence ``\textit{Sam loves Jess more than Jason}'' may mean that Sam loves Jess more than he loves Jason, or that Sam loves Jess more than Jason loves Jess \cite{sennet-ambiguity-23}. 

\subsection{Collective/Distributive Ambiguity}
Some sentences containing plural expressions may be ambiguous between a collective and a distributive reading \cite{sennet-ambiguity-23, dotlacil-2021-collective}. For example, the sentence ``\textit{The students wrote a paper}'' could mean that the students wrote a paper together (collective reading), or that each student wrote a paper separately (distributive reading).

\subsection{Implicative Ambiguity}
It may be ambiguous whether a sentence carries certain implicatures. Sentences are often understood to have pragmatic, implied meanings beyond what is literally said \cite{kearns-2000-semantics, Levinson-1983-pragmatics}. Consider the sentence ``\textit{some gems in this box are fake}.'' The quantifier \emph{some} likely means ``\textit{some and not all}'' since a stronger quantifier wasn’t used instead. However, suppose someone says that sentence after examining a few of the gems and finding each of them to be fake---then \emph{some} may indeed mean ``\textit{some, and perhaps all}.''

\subsection{Presuppositional Ambiguity}
The presuppositions carried by a sentence may be ambiguous \cite{sennet-ambiguity-23}. The word \textit{too} in ``\textit{Jane left early too}'' might trigger the presupposition that Jane was not the only one to leave early (e.g., ``\textit{Robert left early. Jane left early too.}''), or that Jane did something else in addition to leaving early (e.g., ``\textit{Jane arrived early. Jane left early too.}'').

\subsection{Idiomatic Ambiguity}
Idiomatic ambiguity occurs when a sequence of words can be interpreted as an idiom, but a literal (word by word) reading is also possible. For example, the expression ``\textit{kick the bucket}'' typically is a euphemism for dying. A literal reading, about hitting a container with one's foot, would be less common, but may be appropriate in some contexts \cite{chafe-1968-idiom}.

\subsection{Coreferential Ambiguity}
Coreferential ambiguity occurs when a pronoun’s reference is unclear. In the sentence ``\textit{Abby told Brittney that she upset Courtney},'' \emph{she} could refer to either Abby or Brittney \cite{yuan-etal-2023-ambicoref}.

\subsection{Generic/Non-Generic Ambiguity}
Both a generic and a non-generic reading may be possible for some sentences. For instance, ``\textit{dinosaurs ate kelp}'' could describe a general characteristic of dinosaurs, or refer to a particular event of some dinosaurs eating kelp \cite{sennet-ambiguity-23, carlson-1977-bareplural}. Similarly, ``\textit{John ate breakfast with a gold fork}'' could be about how John prefers to eat breakfast, or about a specific breakfast that John had \cite{sennet-ambiguity-23}.

\subsection{Type/Token Ambiguity}
Terms may be ambiguous between a type and a token reading \cite{wetzel-types-tokens-2018, sennet-ambiguity-23}. Consider ``\textit{I paid for the same car}'': if \emph{car} is a type, the speaker means that they bought the kind of car that the listener has. If \emph{car} is a token, the speaker means that they paid for a car twice. \cite{sennet-ambiguity-23}.

\section{Conclusion}
Our taxonomy aims to contribute to the understanding of ambiguity handling in NLP. A close look at the types of ambiguity found in natural language reveals diverse phenomena that may pose different challenges for language models. Developing more nuanced benchmarks and studying models' behavior for each ambiguity type will provide greater insight into NLP systems' abilities to detect and resolve ambiguity.

% \subsection{References}
% this appears at the end
\nocite{}

% Entries for the entire Anthology, followed by custom entries
\bibliography{anthology,custom}
\bibliographystyle{acl_natbib}

\newpage
\onecolumn

\appendix

\section{Examples from \ambient}
\label{sec:appendix}

\begin{table}[b!]
    \resizebox{\textwidth}{!}{%
    \begin{tabular}{clll}
        \toprule
        \textbf{Type} & \textbf{Example} & \textbf{Disambiguation 1} & \textbf{Disambiguation 2} \\\midrule

        \rotatebox[origin=c]{90}{\makecell[c]{\textit{Lexical}}}
        &
        \makecell*[{{p{7.8cm}}}]{P: The \ul{speaker} is at the front of the room.\\
        H: There is a loudspeaker in the room.\\
        $\lbag\textcolor{ent}{\texttt{\textbf{ENTAIL}}}, \textcolor{neu}{\texttt{\textbf{NEUTRAL}}}\rbag$
        $\>$}
        & \makecell[{{p{4.8cm}}}]{P: The \ul{loudspeaker} is at the front of the room.\\
        $\textcolor{ent}{\texttt{\textbf{ENTAIL}}}$
        $\>$}
        &\makecell*[{{p{4.8cm}}}]{P: The \ul{person who is speaking} is at the front of the room.\\
        $\textcolor{neu}{\texttt{\textbf{NEUTRAL}}}$
        $\>$}
        \\\midrule

        \rotatebox[origin=c]{90}{\makecell[c]{\textit{Syntactic}}}
        &
        \makecell*[{{p{7.8cm}}}]{P: He's drawing all over the bus with graffiti.\\
        H: He's drawing \ul{on the bus}. \\
        $\lbag\textcolor{neu}{\texttt{\textbf{NEUTRAL}}}, \textcolor{ent}{\texttt{\textbf{ENTAIL}}}\rbag$
        $\>$}
        & \makecell[{{p{4.8cm}}}]{H: \ul{He's on the bus}, drawing.\\
        $\textcolor{neu}{\texttt{\textbf{NEUTRAL}}}$
        $\>$}
        &\makecell*[{{p{4.8cm}}}]{H: He's drawing \ul{on the surface of the bus}.\\
        $\textcolor{ent}{\texttt{\textbf{ENTAIL}}}$
        $\>$}
        \\\midrule

        \rotatebox[origin=c]{90}{\makecell[c]{\textit{Scopal}}} &
        \makecell*[{{p{7.8cm}}}]{P: He wants to attend \ul{a school in New York}.\\
        H: The location of a school does not matter for him. \\
        $\lbag\textcolor{neu}{\texttt{\textbf{NEUTRAL}}}, \textcolor{con}{\texttt{\textbf{CONTRADICTION}}}\rbag$
        $\>$}
        & \makecell[{{p{4.8cm}}}]{P: \ul{There is a school in New York} that he wants to attend.\\
        $\textcolor{neu}{\texttt{\textbf{NEUTRAL}}}$
        $\>$}
        &\makecell*[{{p{4.8cm}}}]{P: He wants \ul{to be in New York for school}. \\
        $\textcolor{con}{\texttt{\textbf{CONTRADICTION}}}$
        $\>$}
        \\\midrule

        \rotatebox[origin=c]{90}{\makecell[c]{\textit{Elliptical}}} &
        \makecell*[{{p{7.8cm}}}]{P: Calvin will honor his father and Otto \ul{will too}.\\
        H: Otto will not honor Calvin's father. \\
        $\lbag\textcolor{con}{\texttt{\textbf{CONTRADICTION}}}, \textcolor{neu}{\texttt{\textbf{NEUTRAL}}}\rbag$
        $\>$}
        & \makecell[{{p{4.8cm}}}]{P: Calvin will honor his father and Otto \ul{will honor Calvin's father too}.\\
        $\textcolor{con}{\texttt{\textbf{CONTRADICTION}}}$
        $\>$}
        &\makecell*[{{p{4.8cm}}}]{P: Calvin will honor his father and Otto \ul{will honor his own father too}. \\
        $\textcolor{neu}{\texttt{\textbf{NEUTRAL}}}$
        $\>$}
        \\\midrule

        \rotatebox[origin=c]{90}{\makecell[c]{\textit{Collective/} \\\textit{Distributive}}} &
        \makecell*[{{p{7.8cm}}}]{P: Jenny and Zoe \ul{solved the puzzle}.\\
        H: They solved it together. \\
        $\lbag\textcolor{con}{\texttt{\textbf{CONTRADICTION}}}, \textcolor{ent}{\texttt{\textbf{ENTAIL}}}\rbag$
        $\>$}
        & \makecell[{{p{4.8cm}}}]{P: Jenny and Zoe \ul{each solved the puzzle}. \\
        $\textcolor{con}{\texttt{\textbf{CONTRADICTION}}}$
        $\>$}
        &\makecell*[{{p{4.8cm}}}]{P: Jenny and Zoe \ul{solved the puzzle together}. \\
        $\textcolor{ent}{\texttt{\textbf{ENTAIL}}}$
        $\>$}
        \\\midrule

        \rotatebox[origin=c]{90}{\makecell[c]{\textit{Implicative}}} &
        \makecell*[{{p{7.8cm}}}]{P: Carolyn had talked to \ul{two} senators. \\
        H: Carolyn had talked to three senators. \\
        $\lbag\textcolor{con}{\texttt{\textbf{CONTRADICTION}}}, \textcolor{neu}{\texttt{\textbf{NEUTRAL}}}\rbag$
        $\>$}
        & \makecell[{{p{4.8cm}}}]{P: Carolyn had talked to \ul{exactly two} senators. \\
        $\textcolor{con}{\texttt{\textbf{CONTRADICTION}}}$
        $\>$}
        &\makecell*[{{p{4.8cm}}}]{P: Carolyn had talked to \ul{at least two} senators. \\
        $\textcolor{neu}{\texttt{\textbf{NEUTRAL}}}$
        $\>$}
        \\\midrule

        \rotatebox[origin=c]{90}{\makecell[c]{\textit{Presuppositional}}} &
        \makecell*[{{p{7.8cm}}}]{P: The new software is \ul{also} available in a Spanish-language version. \\
        H: This is the only software available in a Spanish-language version. \\
        $\lbag\textcolor{con}{\texttt{\textbf{CONTRADICTION}}}, \textcolor{neu}{\texttt{\textbf{NEUTRAL}}}\rbag$
        $\>$}
        & \makecell[{{p{4.8cm}}}]{P: \ul{In addition to older software}, the new software is also available in a Spanish-language version. \\
        $\textcolor{con}{\texttt{\textbf{CONTRADICTION}}}$
        $\>$}
        &\makecell*[{{p{4.8cm}}}]{P: \ul{In addition to other languages}, the new software is also available in a Spanish-language version. \\
        $\textcolor{neu}{\texttt{\textbf{NEUTRAL}}}$
        $\>$}
        \\\midrule

        \rotatebox[origin=c]{90}{\makecell[c]{\textit{Idiomatic}}} &
        \makecell*[{{p{7.8cm}}}]{P: He didn't see \ul{the big picture}. \\
        H: He was missing the point. \\
        $\lbag \textcolor{neu}{\texttt{\textbf{NEUTRAL}}}, \textcolor{ent}{\texttt{\textbf{ENTAIL}}} \rbag$
        $\>$}
        & \makecell[{{p{4.8cm}}}]{P: He didn't see \ul{the physical big picture}. \\
        $\textcolor{neu}{\texttt{\textbf{NEUTRAL}}}$
        $\>$}
        &\makecell*[{{p{4.8cm}}}]{P: He didn't see \ul{the metaphorical big picture}. \\
        $\textcolor{ent}{\texttt{\textbf{ENTAIL}}}$
        $\>$}
        \\\midrule

        \rotatebox[origin=c]{90}{\makecell[c]{\textit{Coreferential}}} &
        \makecell*[{{p{7.8cm}}}]{P: My roommate and I met the lawyer for coffee, but \ul{she} became ill and had to leave. \\
        H: The lawyer left. \\
        $\lbag \textcolor{ent}{\texttt{\textbf{ENTAIL}}}, \textcolor{neu}{\texttt{\textbf{NEUTRAL}}} \rbag$
        $\>$}
        & \makecell[{{p{4.8cm}}}]{P: My roommate and I met the lawyer for coffee, but \ul{the lawyer} became ill and had to leave. \\
        $\textcolor{ent}{\texttt{\textbf{ENTAIL}}}$
        $\>$}
        &\makecell*[{{p{4.8cm}}}]{P: My roommate and I met the lawyer for coffee, but \ul{my roommate} became ill and had to leave. \\
        $\textcolor{neu}{\texttt{\textbf{NEUTRAL}}}$
        $\>$}
        \\\midrule

        \rotatebox[origin=c]{90}{\makecell[c]{\textit{Generic/} \\\textit{Non-Generic}}} &
        \makecell*[{{p{7.8cm}}}]{P: If an athlete uses a banned substance, they will be disqualified from the competition. \\
        H: \ul{The athlete} will not be disqualified from the competition. \\
        $\lbag \textcolor{neu}{\texttt{\textbf{NEUTRAL}}}, \textcolor{con}{\texttt{\textbf{CONTRADICTION}}} \rbag$
        $\>$}
        & \makecell[{{p{4.8cm}}}]{H: \ul{The athlete (a specific person)} will not be disqualified from the competition. \\
        $\textcolor{neu}{\texttt{\textbf{NEUTRAL}}}$
        $\>$}
        &\makecell*[{{p{4.8cm}}}]{H: \ul{The athlete (as a generic)} will not be disqualified from the competition. \\
        $\textcolor{con}{\texttt{\textbf{CONTRADICTION}}}$
        $\>$}
        \\\midrule

        \rotatebox[origin=c]{90}{\makecell[c]{\textit{Type/Token}}} &
        \makecell*[{{p{7.8cm}}}]{P: You should visit Norway in \ul{the summer}. \\
        H: Summer is a good season to visit Norway. \\
        $\lbag \textcolor{neu}{\texttt{\textbf{NEUTRAL}}}, \textcolor{ent}{\texttt{\textbf{ENTAIL}}} \rbag$
        $\>$}
        & \makecell[{{p{4.8cm}}}]{P: You should visit Norway \ul{the coming summer}. \\
        $\textcolor{neu}{\texttt{\textbf{NEUTRAL}}}$
        $\>$}
        &\makecell*[{{p{4.8cm}}}]{P: You should visit Norway in \ul{the summer season}. \\
        $\textcolor{ent}{\texttt{\textbf{ENTAIL}}}$
        $\>$}
        
        % \rotatebox[origin=c]{90}{\makecell[c]{\textit{CATEGORY}}} &
        % \makecell*[{{p{7.8cm}}}]{P: PREMISE \\
        % H: HYPOTHESIS \\
        % $\lbag COLOR1, COLOR2 \rbag$
        % $\>$}
        % & \makecell[{{p{4.8cm}}}]{P: DISAMBIGUATION1 \\
        % $COLOR1$
        % $\>$}
        % &\makecell*[{{p{4.8cm}}}]{P: DISAMBIGUATION2 \\
        % $COLOR2$
        % $\>$}
        % \\\midrule

% \textcolor{ent}{\texttt{\textbf{ENTAIL}}}
% \textcolor{con}{\texttt{\textbf{CONTRADICTION}}}
% \textcolor{neu}{\texttt{\textbf{NEUTRAL}}}
        
        \\\bottomrule
    \end{tabular}}
    \caption{Ambiguous examples in \ambient\ categorized by ambiguity type according to our taxonomy. Each example, formatted as a natural language inference task, contains some ambiguity in the premise or the hypothesis. Instead of a single label, a set of $\lbag$\texttt{\textbf{GOLD LABELS}}$\rbag$ is given for each example---different labels result from different disambiguations.}
    \label{tab:examples}
    \vspace{-10pt}
\end{table}

\end{document}